\documentclass{article}
\usepackage{imav}
\usepackage{times}
\usepackage{graphicx}
\usepackage{amsmath, amssymb}
\usepackage{booktabs}
\usepackage{wasysym}
\usepackage{orcidlink}
\usepackage{siunitx}
\usepackage{url}

\title{Flight-Ready LiDAR-Inertial Odometry for Embedded Drone Platforms}
\author{Alvaro J. Gaona~\orcidlink{0009-0003-4967-4444}, David Perez-Saura~\orcidlink{0000-0003-2571-3165}, Francisco J. Anguita~\orcidlink{0009-0001-3017-9406}, and Pascual Campoy~\orcidlink{0000-0002-9894-2009} \\ Computer Vision and Aerial Robotics Group at Centre for Automation and Robotics C.A.R. (UPM-CSIC), 
Universidad Politécnica de Madrid (CVAR-UPM), Calle Jose Gutierrez Abascal 2, 28006 Madrid, Spain}

\begin{document}

\maketitle
\thispagestyle{empty}

\begin{abstract}
Open-source LiDAR–inertial odometry (LIO) systems have achieved remarkable benchmark accuracy, yet current state-of-the-art implementations are primarily optimized for evaluation performance rather than the requirements of real-time closed-loop aerial control. When deployed onboard UAVs, this can introduce limitations that degrade flight performance. In this work, we identify five architectural deficiencies in a representative tightly coupled IESKF-based LIO implementation: odometry publishing tied to the LiDAR rate (10 Hz instead of the IMU’s 200 Hz), missing velocity outputs, execution bottlenecks that block IMU processing, mutex contention, and synchronization race conditions. We introduce corresponding modifications including IMU-rate forward propagation, direct body-frame velocity publishing, SLERP-based smoothing, dual-executor isolation, and explicit synchronization protection. The resulting system increases odometry output from ~10 Hz to a stable 200 Hz, provides a complete Twist state at every IMU sample, and preserves continuity during transient LiDAR loss. Experiments on a Livox Mid-360 / Pixhawk 4 Mini autonomous UAV with motion-capture ground truth validate the approach. Since the underlying estimator (IESKF + ikd-Tree) remains unchanged, the proposed improvements can be directly applied to FAST-LIO2-derived implementations.
\end{abstract}

\section{Introduction}\label{section:introduction}

\begin{figure}[t]
\centering
\includegraphics[width=.85\columnwidth, trim={6cm 0 6cm 0}, clip]{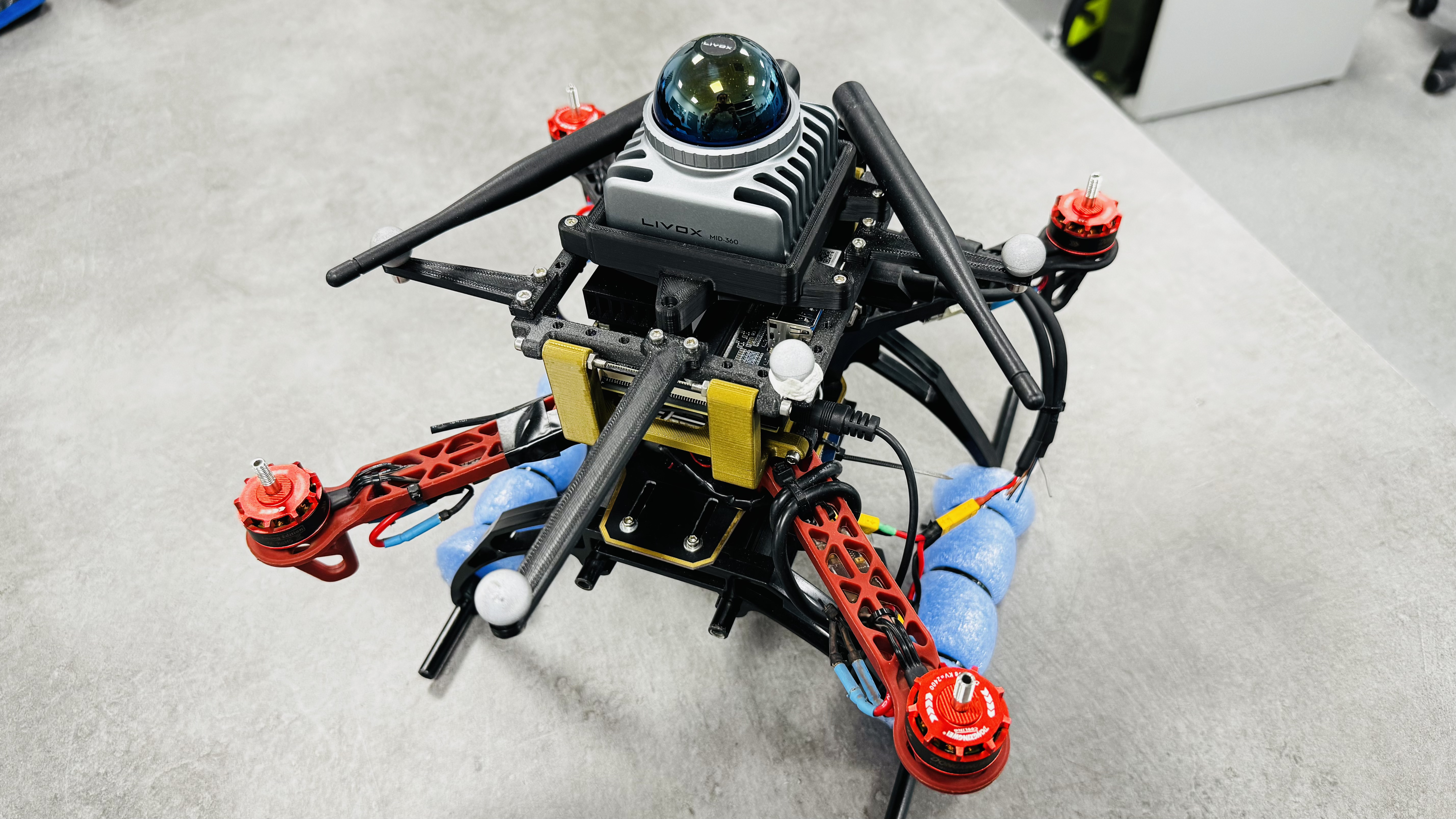}
\caption{Quadrotor platform used for the validation flights: Livox Mid-360 LiDAR on top, Pixhawk 4 Mini autopilot and Jetson Orin NX onboard computer below, reflective mocap markers on the prop guards. Flight footage: \url{https://vimeo.com/1192528650?share=copy&fl=sv&fe=ci}.}
\label{figure:drone}
\end{figure}

Tightly-coupled LiDAR-inertial odometry has matured rapidly over the last five years, with implementations such as FAST-LIO2~\cite{xu2022fastlio2}, LIO-SAM~\cite{shan2020liosam}, Point-LIO~\cite{he2023pointlio}, and DLIO~\cite{chen2023dlio} routinely reporting centimeter-level accuracy on standard benchmark datasets. These implementations are open-source, well-documented, and widely cited; a roboticist building an aerial-mapping system has every reason to start from one of them.

A roboticist building an autonomous \emph{flight} system, however, quickly discovers that benchmark accuracy and flight-readiness are not the same thing. A flight controller imposes requirements that benchmark evaluations do not exercise: pose feedback at hundreds of Hertz to keep the inner control loop stable, linear and angular velocity estimates at the same rate for damping and feedforward, low-noise signals that do not excite actuator oscillations, and uninterrupted continuity even during transient sensor degradation. An LIO implementation that meets none of these requirements out of the box is not unusable; it is simply not yet ready for flight.

This paper documents the modifications required to take a representative open-source IESKF-based LIO implementation from benchmark-tuned to flight-ready.\footnote{Source code: \url{https://github.com/alvgaona/fr-lio}.} The five contributions are:
\begin{enumerate}
    \item A forward-propagation scheme that publishes pose and velocity at every IMU sample (200\,Hz) using the most recent IESKF-corrected state as an anchor.
    \item Body-frame linear and angular velocity derived directly from the filter state, populating the previously-empty field of the odometry message.
    \item A first-order EMA filter on velocity and SLERP-based filtering on orientation, with cutoff frequencies chosen to remove integration noise without obscuring drone-bandwidth maneuvers.
    \item A dual-executor architecture that isolates the IMU callback from scan-processing latency, eliminating rate-degradation under load.
    \item An identification and fix of a race condition in the IMU/LiDAR synchronization function that emerges only under multi-threaded execution.
\end{enumerate}

These modifications are purely architectural: the underlying IESKF state propagation, point-to-plane registration, and ikd-Tree map remain unchanged. They apply to any FAST-LIO2-derived implementation and to any IESKF-based system where the filter correction operates at a rate slower than the IMU.

The paper is organized as follows. Section~\ref{section:related} surveys representative open-source LIO implementations and characterizes the flight-readiness gap each leaves open. Section~\ref{section:background} reviews the publishing pipeline of the baseline implementation and identifies the limitations that motivate each modification. Section~\ref{section:propagation} presents the IMU-rate forward propagation and body-frame velocity derivation. Section~\ref{section:threading} describes the dual-executor architecture, mutex-contention reduction, and synchronization race-condition fix. Section~\ref{section:robustness} covers EMA/SLERP filtering and chained-propagation fallback for LiDAR-interruption robustness. Section~\ref{section:validation} reports flight-test validation on a Livox Mid-360 / Pixhawk 4 Mini platform with motion-capture ground truth. Section~\ref{section:conclusion} concludes.

\section{Related Work}\label{section:related}

Modern LIO implementations have converged on two architectural patterns: tightly-coupled iterated error-state extended Kalman filters (IESKF) with incremental map structures, and tightly-coupled factor graphs with incremental smoothing back-ends. Both deliver centimeter-accurate trajectories on benchmark datasets~\cite{xu2022fastlio2,shan2020liosam,he2023pointlio,chen2023dlio}; neither was designed with the requirements of an in-flight controller in mind. We briefly review the four most widely-used representatives and characterize the flight-readiness gap each leaves open.

\textbf{FAST-LIO2}~\cite{xu2022fastlio2} is the canonical IESKF+\textit{ikd}-Tree implementation. The state is propagated forward by IMU integration over each scan interval, then corrected by a single IESKF iteration loop with point-to-plane registration against the \textit{ikd}-Tree map. Publishing happens once per scan, at the end of the IESKF correction. The default ROS\,1/2 deployment uses a single-threaded executor, so all callbacks (IMU buffering, LiDAR buffering, IESKF processing, publishing) serialize on one thread. The Twist field of the published \texttt{Odometry} message is left at zero. None of these choices is wrong for offline mapping or benchmark evaluation; all four limit flight-readiness directly.

\textbf{LIO-SAM}~\cite{shan2020liosam} replaces the IESKF with a tightly-coupled factor graph optimized incrementally via iSAM2. It supports GPS and loop-closure factors, which makes it well-suited to outdoor mapping. Publishing rate is again tied to the keyframe selection (typically 1--10\,Hz), and the published velocity is derived from preintegrated IMU factors only at keyframe boundaries. The factor-graph back-end is computationally heavier than an IESKF and the per-keyframe variability makes the publishing rate unpredictable, both of which compound the flight-control problem rather than relieve it.

\textbf{Point-LIO}~\cite{he2023pointlio} addresses high-bandwidth motion by treating each LiDAR point as an individual measurement update rather than batching points into accumulated scans. This raises the IESKF correction rate substantially (the authors report up to kilohertz output) and is well-suited to aggressive motions where intra-scan distortion would otherwise dominate. The output frequency is high, but the architecture inherits FAST-LIO2's single-executor concurrency model; under sustained heavy load the same thread still serializes IMU, LiDAR, and publishing. Point-LIO closes one of the four flight-readiness gaps (rate) without addressing the others (velocity availability, executor isolation, race-condition robustness).

\textbf{DLIO}~\cite{chen2023dlio} is a recent lightweight LIO that adds continuous-time motion correction to deskew scans during integration, packaged with a clean ROS\,2 implementation. The lightweight design makes it attractive for embedded deployment, and the ROS\,2 native architecture supports multi-threaded executors in principle. However, the published implementation does not by default isolate the IMU callback to a dedicated executor, does not populate the Twist field with body-frame velocities, and inherits the same publishing-coupled-to-LiDAR pattern as FAST-LIO2. The architectural improvements are at the estimation layer, not at the deployment layer.

\textbf{Summary.} Across all four representative implementations, the flight-readiness gap manifests as the same four issues: publishing rate coupled to (or capped by) the LiDAR rate, missing velocity fields, single-threaded execution that blocks the IMU callback under load, and an absence of any mechanism to maintain odometry continuity during transient LiDAR loss. Table~\ref{table:related} summarizes which axes each implementation addresses out of the box. The modifications presented in this paper close all four gaps simultaneously and apply directly to any of these implementations, since they intervene at the publishing layer rather than at the estimation layer; the underlying choice of IESKF vs.\ factor graph, batched scans vs.\ per-point updates, or discrete vs.\ continuous-time motion correction is independent of the architectural fix. Visual-inertial state estimators such as VINS-Mono~\cite{qin2018vinsmono} and OpenVINS~\cite{geneva2020openvins} already publish at IMU rate by propagating the corrected state between visual updates; our contribution can be read as adapting that publishing pattern to LIO.

\begin{table}[hbt]
\begin{center}
\footnotesize
\setlength{\tabcolsep}{4pt}
\begin{tabular}{lcccc}
\toprule
                                    & 200\,Hz       & Velocity       & Isolated       & LiDAR-loss \\
                                    & rate          & output         & IMU thread     & resilience \\
\midrule
FAST-LIO2~\cite{xu2022fastlio2}     & no            & no             & no             & no \\
LIO-SAM~\cite{shan2020liosam}       & no            & partial        & no             & no \\
Point-LIO~\cite{he2023pointlio}     & yes           & no             & no             & no \\
DLIO~\cite{chen2023dlio}            & no            & no             & no             & no \\
\midrule
\textbf{This work}                  & \textbf{yes}  & \textbf{yes}   & \textbf{yes}   & \textbf{yes} \\
\bottomrule
\end{tabular}
\caption{Flight-readiness axes addressed by representative open-source LIO implementations. ``Velocity output'' = body-frame linear and angular velocities published alongside the pose. ``Isolated IMU thread'' = IMU sampling runs on a dedicated thread, decoupled from scan-processing latency. ``LiDAR-loss resilience'' = mechanism to maintain odometry continuity when LiDAR scans stop arriving for a transient period.}
\label{table:related}
\end{center}
\end{table}

\section{Background and Limitations}\label{section:background}

The baseline implementation publishes odometry inside the LiDAR scan-processing loop. Each cycle accumulates a complete scan, executes IMU forward propagation across the scan interval, performs IESKF iterations with point-to-plane registration, updates the ikd-Tree map, and publishes the resulting pose. The publishing rate is therefore tied to the LiDAR scan-arrival rate.

For the Livox Mid-360 sensor at 100\,Hz, an accumulator collecting 10 packets per IESKF cycle yields a $\sim$10\,Hz publishing rate. At a flight speed of 2\,m/s, the controller receives a new pose every 0.2\,m of travel and must extrapolate between updates---an extrapolation that is unsafe for tight closed-loop control. Three structural issues underlie this:

\textbf{Publishing rate coupled to LiDAR rate.} When scan processing exceeds the inter-scan period, for example during heavy ikd-Tree insertions, the publishing rate degrades below 10\,Hz. The rate is therefore not a guarantee but a best-case.

\textbf{Absence of velocities.} The published odometry message carries only the pose; the velocity component is left at zero. The flight controller must numerically differentiate position to obtain velocity, amplifying high-frequency noise and introducing at least one sampling period of lag.

\textbf{Single-threaded execution.} The baseline serializes IMU buffering, LiDAR buffering, IESKF processing, and publishing on a single thread. While the IESKF processes a scan, the IMU samples are queued. When publishing happens at the end of scan processing this serialization is harmless; moving the publication to the IMU sample rate to achieve 200\,Hz makes blocking the dominant bottleneck.

\section{IMU-Rate Forward Propagation}\label{section:propagation}

\subsection{Operating Principle}

The IESKF correction occurs at the accumulated LiDAR rate ($\sim$10\,Hz). Between corrections, each new IMU sample propagates the state forward using the inertial dynamics. The propagation does not replace the filter; it uses the most recent corrected state as an \emph{anchor} and applies single-step integration with the current IMU measurement.

Let $t_a$ denote the time of the last IESKF correction and $(\mathbf{R}_a, \mathbf{p}_a, \mathbf{v}_a, \mathbf{b}_{g,a}, \mathbf{b}_{a,a}, \mathbf{g}_a)$ the corrected anchor state. For an IMU sample at time $t > t_a$ with raw measurements $\boldsymbol{\omega}_m$ and $\mathbf{a}_m$, the bias-corrected angular velocity and body-frame acceleration are
\begin{align}
    \boldsymbol{\omega} &= \boldsymbol{\omega}_m - \mathbf{b}_{g,a}, \quad \mathbf{a}_b = \mathbf{a}_m - \mathbf{b}_{a,a},\label{eq:bias-correct}
\end{align}
and the world-frame acceleration is $\mathbf{a}_w = \mathbf{R}_a \mathbf{a}_b + \mathbf{g}_a$. With $\Delta t = t - t_a$, the propagated state is
\begin{align}
    \mathbf{p}(t) &= \mathbf{p}_a + \mathbf{v}_a \Delta t + \tfrac{1}{2} \mathbf{a}_w \Delta t^2, \label{eq:prop-pos} \\
    \mathbf{v}(t) &= \mathbf{v}_a + \mathbf{a}_w \Delta t, \label{eq:prop-vel} \\
    \mathbf{R}(t) &= \mathbf{R}_a \, \mathrm{Exp}(\boldsymbol{\omega} \Delta t),\label{eq:prop-rot}
\end{align}
where $\mathrm{Exp}: \mathbb{R}^3 \to \mathrm{SO}(3)$ is the SO(3) exponential map. Equation~\eqref{eq:prop-rot} is the zeroth-order discretization, treating $\boldsymbol{\omega}$ as constant over $\Delta t$; this is appropriate at the 5\,ms IMU period. The integration also assumes that $\mathbf{R}_a$, the biases, and gravity remain constant over $\Delta t$. For $\Delta t < 0.1$\,s (the nominal $10$\,Hz IESKF rate), the resulting error is on the order of centimeters in position and fractions of a degree in orientation~\cite{sola2017quaternion}---negligible relative to the controller's resolution.

\subsection{Anchor Update}

Each IESKF correction copies its output state to the propagation anchor under mutex protection. The anchor stores position, velocity, rotation, gyroscope and accelerometer biases, gravity, and the timestamp of the corresponding scan end. The next IMU sample resets $\Delta t$ to a small value, so propagation never accumulates error beyond the inter-correction interval.

\subsection{Body-Frame Velocities}

The IESKF state vector includes velocity in the world frame, $\mathbf{v} \in \mathbb{R}^3$. The flight controller requires velocity in the body frame, obtained via the inverse rotation:
\begin{equation}
    \mathbf{v}_b(t) = \mathbf{R}(t)^\top \mathbf{v}(t).\label{eq:vel-body}
\end{equation}
The angular velocity in the body frame is the bias-corrected gyroscope measurement, available directly from~\eqref{eq:bias-correct}. Both quantities populate the previously-empty Twist field of the published \texttt{Odometry} message.

\section{Threading and Concurrency}\label{section:threading}

\subsection{Dual-Executor Isolation}

To publish at 200\,Hz, the IMU callback must execute without delay at every sample. A single-threaded executor cannot satisfy this when scan processing takes tens of milliseconds (typical during heavy ikd-Tree insertions): under load, the IMU callback is starved for the duration of every IESKF cycle and the effective publishing rate degrades to the LiDAR rate.

The fix uses two independent single-threaded executors. The IMU callback is registered to an exclusive callback group assigned to a dedicated executor that runs on its own thread; the LiDAR callback, IESKF timer callback, and point-cloud publishers are assigned to the main executor. The IMU thread is therefore decoupled from scan-processing time and contends only for shared mutexes, not for execution slots.

\subsection{Mutex Contention from In-Buffer Preprocessing}

Even with separate executors, the IMU thread can be blocked if it requires a mutex held by the main thread. In the baseline, the LiDAR callback acquires the buffer mutex \emph{before} performing scan preprocessing (point filtering, format conversion), an operation that takes several milliseconds.

The fix is to perform preprocessing outside the mutex-protected section. The mutex is acquired only for the buffer insertion of the preprocessed scan, an operation that takes microseconds. This reduces mutex hold time by approximately three orders of magnitude.

\subsection{Race Condition in Synchronization}

The packet synchronization function pairs LiDAR scans with IMU measurements to form IESKF inputs. In the baseline, this function accesses shared buffers without mutex protection. Under single-threaded execution this is safe; under the dual-executor architecture, concurrent writes from the IMU callback and reads from the synchronization function constitute a data race that occasionally produces malformed input pairs and silently degrades estimator accuracy.

The fix protects the synchronization function with the same mutex that guards the buffers. The mutex critical section is short (single-pass over both buffers), and the operation does not appear in any benchmark trace because it is masked by IESKF processing.

\subsection{Path Message Serialization}

A secondary bottleneck arises from publishing the \texttt{nav\_msgs::Path} message at the IMU rate. \texttt{Path} accumulates the full trajectory; at 200\,Hz, after 30\,s of flight the message contains 6000 poses and serialization consumes significant CPU on the IMU thread. The fix throttles \texttt{Path} publication to 10\,Hz; the message is visualization data, not control data, and the rate reduction has no functional effect.

\begin{figure}[hbt]
\centering
\includegraphics[width=.95\columnwidth]{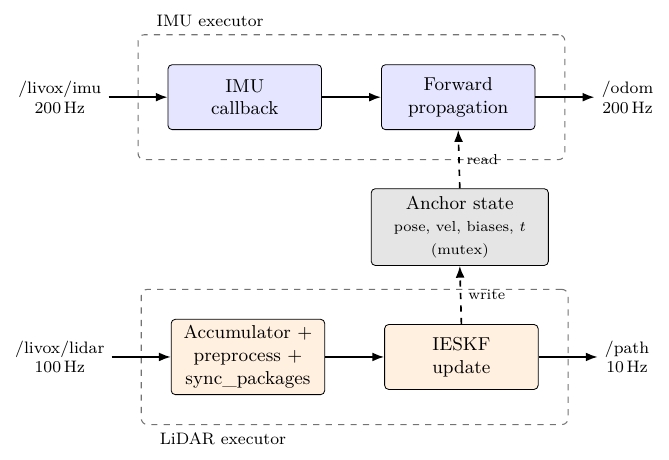}
\caption{Dual-executor threading architecture. The IMU callback runs on its own single-threaded executor (top) and publishes the 200\,Hz \texttt{/odom} stream by forward-propagating from the shared anchor state. LiDAR preprocessing and IESKF correction run on a second executor (bottom) and write the anchor on each successful update. The anchor state (pose, velocity, biases, timestamp) is mutex-protected; the critical section is short and the synchronization function acquires the same mutex.}
\label{figure:architecture}
\end{figure}

\section{Filtering and Robustness}\label{section:robustness}

\subsection{EMA on Velocity and SLERP on Orientation}

Single-step integration of accelerometer measurements transfers high-frequency noise directly to the propagated velocity, producing typical fluctuations of $\sim$0.2\,m/s (and several-m/s peak excursions on the vertical axis, see Figure~\ref{figure:velocity}) that a finely-tuned velocity controller would amplify into actuator oscillations. We apply a first-order exponential moving average (EMA) to the published body-frame velocities:
\begin{equation}
    \bar{\mathbf{v}}_b[k] = \alpha \mathbf{v}_b[k] + (1 - \alpha) \bar{\mathbf{v}}_b[k-1],\label{eq:ema}
\end{equation}
with the same form applied to the body-frame angular velocity. The cutoff frequency at sampling rate $f_s$ is $f_c = (f_s / 2\pi) \ln(1/(1-\alpha))$. For $f_s = 200$\,Hz and $\alpha = 0.05$, $f_c \approx 1.6$\,Hz, which removes accelerometer noise without significantly attenuating drone maneuvers under the precision-hover and indoor-trajectory regimes used in our validation (closed-loop bandwidth $\lesssim 1$\,Hz). Aggressive flight regimes (racing-style trajectories, heavy payload swings) extend the controller bandwidth above this cutoff and require a larger $\alpha$ at the cost of higher noise pass-through.

Quaternion filtering cannot be performed component-wise. We use spherical linear interpolation (SLERP):
\begin{equation}
    \bar{\mathbf{q}}[k] = \mathrm{SLERP}\left(\bar{\mathbf{q}}[k-1], \mathbf{q}[k], \alpha\right),\label{eq:slerp}
\end{equation}
which respects the manifold structure of $S^3$ and reduces to weighted-and-normalized linear interpolation for small angular changes between consecutive samples.

Linear-velocity filtering is applied \emph{after} the rotation to the body frame, not before; filtering in the world frame and then rotating injects gyroscope noise into the otherwise-clean filtered signal.

\subsection{Chained Propagation under LiDAR Interruption}

If LiDAR scans stop arriving (sensor obstruction, packet loss in the DDS network, transient driver failure), the anchor timestamp freezes and $\Delta t$ grows without bound. For large $\Delta t$, the constant-rotation assumption of~\eqref{eq:prop-pos} breaks down and integration error becomes quadratic.

When $\Delta t$ exceeds 0.5\,s, the propagation is \emph{chained}: the propagated state becomes the new anchor for subsequent IMU samples,
\begin{align}
    \mathbf{p}_a, \mathbf{v}_a, \mathbf{R}_a, t_a &\leftarrow \mathbf{p}(t_k), \mathbf{v}(t_k), \mathbf{R}(t_k), t_k.\label{eq:chain}
\end{align}
Biases and gravity are not updated during chaining, since the IESKF estimate remains the best available. Each subsequent integration operates on a small $\Delta t$ (the IMU period, 5\,ms), avoiding quadratic error accumulation.

\subsection{Anchor Timestamp Guard}

When IESKF processing of a scan takes longer than the inter-scan period, the IESKF may attempt to overwrite the anchor with a state whose timestamp is earlier than what chained propagation has already advanced to. This would produce a backward jump in the published odometry. The fix conditions the write of position, velocity, rotation, and timestamp on the new timestamp being later than the current anchor. Biases and gravity are always updated, since they are independent of the time index.

\section{Flight-Test Validation}\label{section:validation}

We validate the modifications on the quadrotor shown in Figure~\ref{figure:drone}: a Livox Mid-360 LiDAR, a Pixhawk 4 Mini autopilot, and a Jetson Orin NX running ROS\,2 (Humble), with a takeoff weight of approximately 1.8\,kg and a 4S 5000\,mAh battery. Validation flights were conducted in the Drone Arena of the Centre for Automation and Robotics (CAR), Universidad Polit\'ecnica de Madrid, operated by the Computer Vision \& Aerial Robotics (CVAR) Group. The arena is a $6 \times 6 \times 12$\,m volume instrumented with 16 OptiTrack motion-capture cameras providing 6-DOF ground-truth pose at \qty{100}{\hertz} with sub-millimeter accuracy.

The flight stack is built on the Aerostack2 framework~\cite{fernandezcortizas2024aerostack2}, which provides the high-level mission management, behaviour layer, and state-machine orchestration. The inner-loop position and attitude controller is a differential-flatness controller~\cite{mellinger2011minimum} that exploits the quadrotor's flatness property to compute attitude and thrust setpoints from desired position trajectories and their derivatives; this controller is the principal consumer of the high-rate odometry and velocity stream produced by the modifications described in this paper. Two autonomous indoor flight sequences are used in what follows. A 58\,s sequence, replayed under identical inputs through both the baseline FAST-LIO and the modified implementation, supports the publishing-rate histogram and the trajectory-accuracy comparison (Figure~\ref{figure:rate}, Table~\ref{table:ate}). A separate 60\,s sequence supports the body-frame velocity tracking analysis (Figure~\ref{figure:velocity}); the same sequence, with a 1.5\,s contiguous block of \texttt{/livox/lidar} messages removed offline before replay through both implementations, is used to exercise the chained-propagation fallback (Figure~\ref{figure:interruption}).

\subsection{Publishing Rate}

\begin{figure}[hbt]
\centering
\includegraphics[width=.95\columnwidth]{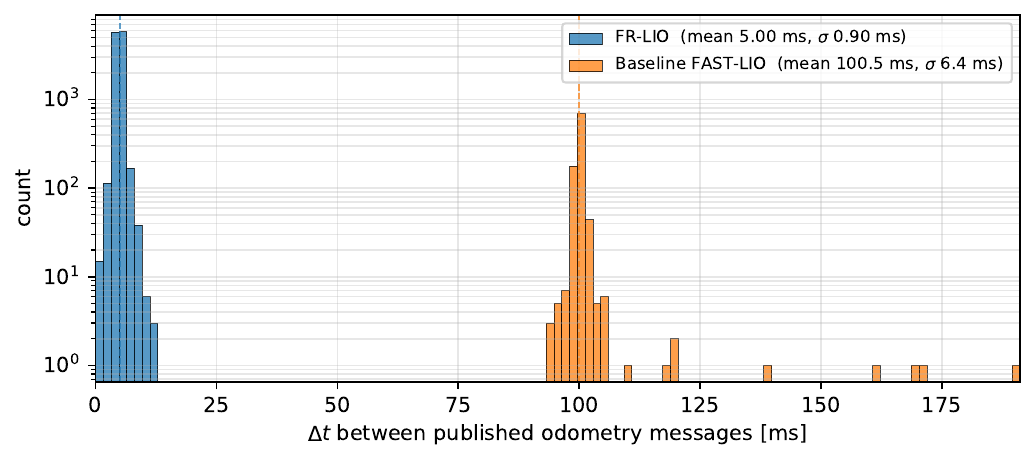}
\caption{Inter-message interval distribution for the published odometry topic on a log-scale y-axis. The modified implementation (blue) concentrates at $\Delta t = 5$\,ms ($f = 200$\,Hz, $\sigma = 0.90$\,ms), while baseline FAST-LIO (orange) publishes at the scan rate near $\Delta t = 100$\,ms ($\sigma = 6.4$\,ms).}
\label{figure:rate}
\end{figure}

Figure~\ref{figure:rate} shows the inter-message interval distribution for the published odometry topic, comparing the baseline against the modified implementation. The modified implementation publishes at a mean interval of 5.00\,ms ($\sigma = 0.90$\,ms, $p_{99} = 6.95$\,ms) — a steady 200\,Hz — while the baseline distribution sits at 100.5\,ms ($\sigma = 6.4$\,ms, $p_{99} = 107.8$\,ms), reflecting its scan-rate publishing.

\subsection{Trajectory Accuracy}

\begin{table}[hbt]
\begin{center}
\begin{tabular}{lcc}
\toprule
Metric & Baseline & Modified \\
\midrule
ATE RMSE (cm)  & 11.7 & 5.2 \\
ATE median (cm) & 4.7  & 3.8 \\
ATE max (cm)   & 32.4 & 28.3 \\
RPE 1\,s (cm)  & 4.9  & 2.4 \\
\bottomrule
\end{tabular}
\caption{Trajectory accuracy against motion-capture ground truth on a 58\,s indoor flight. Each LIO trajectory is rigidly aligned to ground truth via least-squares SE(3) over the first 5\,s of motion (motion onset detected at GT speed $>$\,0.1\,m/s sustained for 0.3\,s).}
\label{table:ate}
\end{center}
\end{table}

Table~\ref{table:ate} compares the published trajectory against motion-capture ground truth. The modified implementation matches the baseline on median ATE within 1\,cm while approximately halving the RPE (1\,s) drift, reflecting the tighter timing of the dual-executor architecture without altering the iterated Kalman update. The ATE max values (28--32\,cm) for both implementations are isolated transients during rapid yaw segments where the IESKF registration is briefly mis-constrained by the Mid-360's reduced lateral overlap; the median and RMSE are the load-bearing comparison numbers.

\subsection{Velocity Tracking}

\begin{figure}[hbt]
\centering
\includegraphics[width=.95\columnwidth]{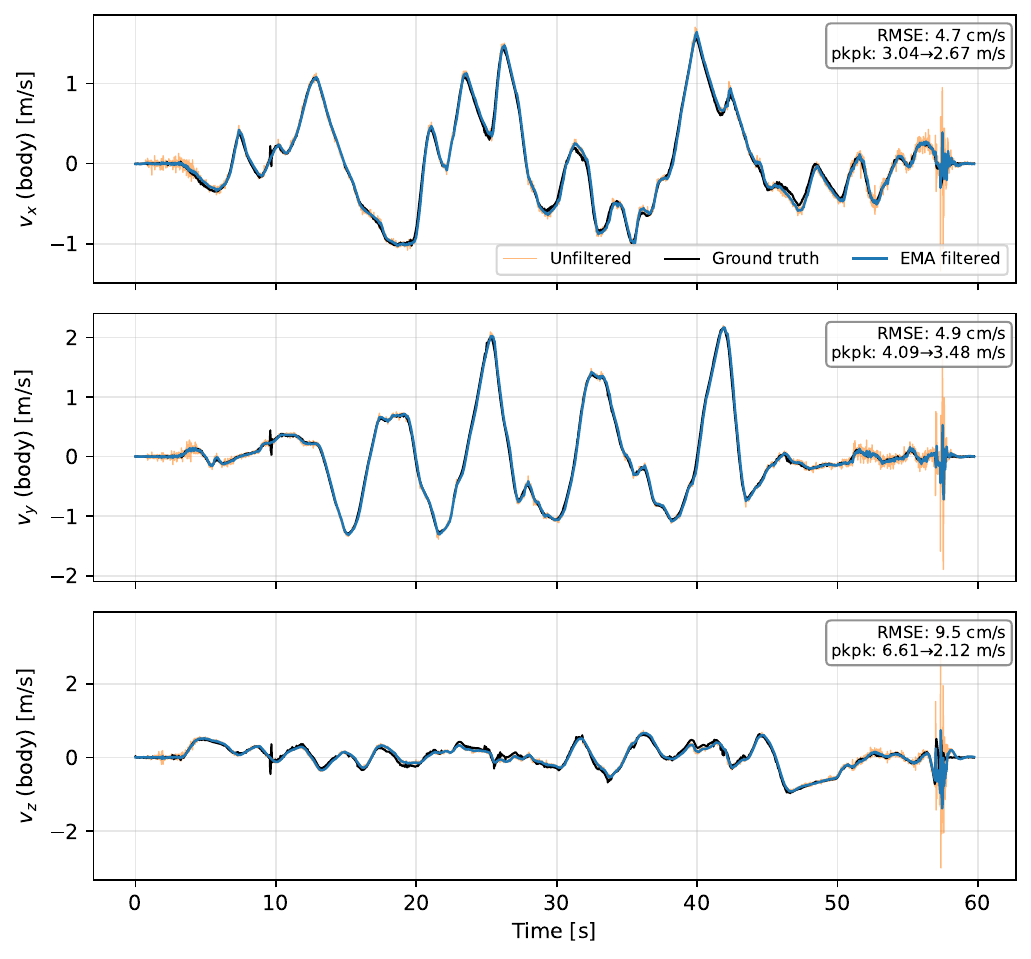}
\caption{Body-frame linear-velocity tracking against motion-capture ground truth. EMA filter ($\alpha = 0.05$, $f_c \approx 1.6$\,Hz) removes accelerometer noise without lagging the controller-bandwidth signal.}
\label{figure:velocity}
\end{figure}

Figure~\ref{figure:velocity} compares published body-frame velocities against ground-truth velocities derived from motion-capture position with a zero-phase Savitzky--Golay smoother (21-tap, order 3). Over the 60\,s flight, the EMA-filtered output tracks ground truth with RMSE 0.05\,m/s on the horizontal axes ($v_x$, $v_y$) and 0.10\,m/s on $v_z$. The high-frequency content removed by the EMA, measured as the standard deviation of the unfiltered--filtered difference, is 5--11\,cm/s per axis. The largest reduction is on $v_z$, where the propagator's peak-to-peak noise drops from 6.6\,m/s to 2.1\,m/s, eliminating the high-bandwidth content that would otherwise excite the thrust loop. The magnitude of the unfiltered $v_z$ noise reflects the operating environment of a body-mounted IMU on a quadrotor: motor harmonics in the 100--300\,Hz range alias into the 200\,Hz sample rate, and even a small misalignment of the estimated gravity vector projects onto $v_z$ when integrated. The EMA at $f_c \approx 1.6$\,Hz attenuates this content without touching the sub-Hz controller-bandwidth signal.

\subsection{LiDAR-Interruption Resilience}

\begin{figure}[hbt]
\centering
\includegraphics[width=.95\columnwidth]{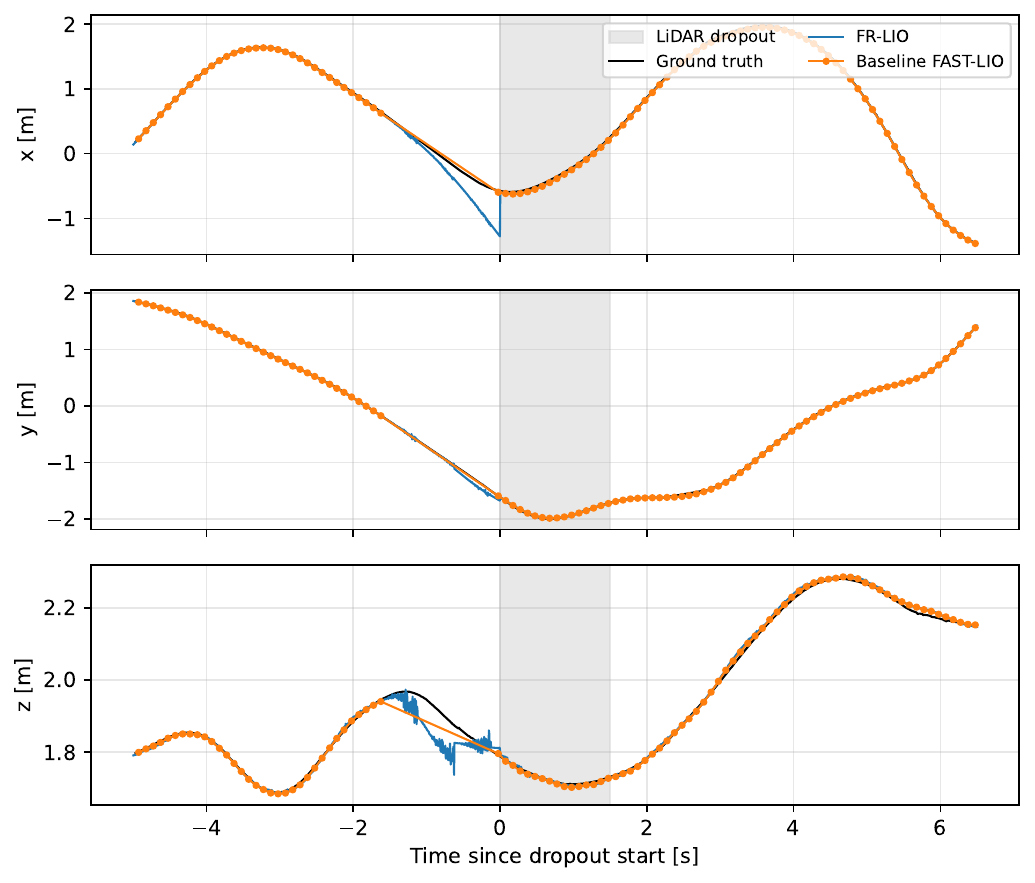}
\caption{Position traces (x, y, z) before, during, and after a 1.5\,s synthetic LiDAR dropout (gray band). The modified implementation (blue) continues publishing through the dropout via chained IMU propagation, accumulating $\sim$0.49\,m of drift in body-frame position; baseline FAST-LIO (orange) freezes at the last EKF state until LiDAR resumes. Both implementations re-converge to ground truth (black) within one update after LiDAR returns.}
\label{figure:interruption}
\end{figure}

To exercise the chained-propagation fallback, we apply the offline 1.5\,s LiDAR-removal described above to the 60\,s sequence used in Figure~\ref{figure:velocity} and replay the resulting bag through both implementations. Figure~\ref{figure:interruption} shows the resulting position traces. The modified implementation continues publishing throughout, accumulating approximately 49\,cm of drift across the dropout. This drift reflects open-loop IMU integration at hover and converges back to ground truth on the next IESKF correction. The baseline implementation freezes at the last EKF state and resumes only when LiDAR returns. A controller running at 200\,Hz under the modified implementation thus sees a smoothly drifting estimate rather than a discontinuity, which is the property that keeps the inner loop stable across the dropout.

\section{Conclusion}\label{section:conclusion}



We presented a set of architectural modifications that transform a benchmark-oriented tightly coupled LiDAR–inertial odometry implementation into a flight-ready system for embedded UAV platforms. Experimental results demonstrate that these changes substantially improve deployment characteristics without altering the underlying estimator architecture. As shown in Figure 3, the proposed system increases odometry output from the baseline scan-coupled rate of approximately 100 ms between updates (~10 Hz) to a stable 5.0 ms interval (200 Hz), while significantly reducing timing variability.
The modifications also preserve—and in some cases improve—estimation quality. Table 2 shows that trajectory accuracy remains comparable to the baseline in median ATE (3.8 cm vs. 4.7 cm), while reducing ATE RMSE from 11.7 cm to 5.2 cm and approximately halving the 1 s relative pose error (4.9 cm to 2.4 cm). Body-frame velocity estimation further achieves tracking errors below 0.1 m/s on all axes while attenuating high-frequency noise, particularly on the vertical channel where peak-to-peak fluctuations decrease from 6.6 m/s to 2.1 m/s (Figure 4).

Finally, Figure 5 demonstrates resilience to transient sensor failures: during a synthetic 1.5 s LiDAR interruption, the proposed approach continues publishing state estimates and limits drift accumulation to approximately 0.49 m, whereas the baseline freezes until measurements resume. Together, these results show that the proposed modifications improve timing, velocity availability, and robustness while preserving estimator accuracy. Because the IESKF + ikd-Tree estimation core remains unchanged, the approach can be directly integrated into FAST-LIO2-derived systems and generalized to other LIO implementations where correction updates operate more slowly than IMU propagation.

\section*{Acknowledgment}

This work has been supported by the project SHEREC “Safe Healthy and Environmental Ship Recycling", Referece: 101136056, funded by European Union under the Horizon Europe Program HORIZON-CL4-2023-HUMAN-01 CNECT.

\bibliographystyle{unsrt}
\bibliography{bibliography}

@article{xu2022fastlio2,
  author  = {Xu, Wei and Cai, Yixi and He, Dongjiao and Lin, Jiarong and Zhang, Fu},
  title   = {{FAST-LIO2}: Fast Direct {LiDAR}-Inertial Odometry},
  journal = {IEEE Transactions on Robotics},
  volume  = {38},
  number  = {4},
  pages   = {2053--2073},
  year    = {2022},
  month   = aug,
  doi     = {10.1109/TRO.2022.3141876}
}

@inproceedings{shan2020liosam,
  author    = {Shan, Tixiao and Englot, Brendan and Meyers, Drew and Wang, Wei and Ratti, Carlo and Rus, Daniela},
  title     = {{LIO-SAM}: Tightly-Coupled Lidar Inertial Odometry via Smoothing and Mapping},
  booktitle = {Proceedings of the IEEE/RSJ International Conference on Intelligent Robots and Systems (IROS)},
  pages     = {5135--5142},
  year      = {2020},
  doi       = {10.1109/IROS45743.2020.9341176}
}

@article{he2023pointlio,
  author  = {He, Dongjiao and Xu, Wei and Chen, Nan and Kong, Fanze and Yuan, Chongjian and Zhang, Fu},
  title   = {{Point-LIO}: Robust High-Bandwidth Light Detection and Ranging Inertial Odometry},
  journal = {Advanced Intelligent Systems},
  volume  = {5},
  number  = {7},
  pages   = {2200459},
  year    = {2023},
  doi     = {10.1002/aisy.202200459}
}

@inproceedings{chen2023dlio,
  author    = {Chen, Kenny and Nemiroff, Ryan and Lopez, Brett T.},
  title     = {Direct {LiDAR}-Inertial Odometry: Lightweight {LIO} with Continuous-Time Motion Correction},
  booktitle = {Proceedings of the IEEE International Conference on Robotics and Automation (ICRA)},
  pages     = {3983--3989},
  year      = {2023},
  doi       = {10.1109/ICRA48891.2023.10160508}
}

@misc{sola2017quaternion,
  author       = {Sol{\`a}, Joan},
  title        = {Quaternion Kinematics for the Error-State {Kalman} Filter},
  year         = {2017},
  eprint       = {1711.02508},
  archivePrefix= {arXiv},
  primaryClass = {cs.RO},
  note         = {arXiv:1711.02508}
}

@misc{fernandezcortizas2024aerostack2,
  title         = {Aerostack2: A Software Framework for Developing Multi-robot Aerial Systems},
  author        = {Miguel Fernandez-Cortizas and Martin Molina and Pedro Arias-Perez and Rafael Perez-Segui and David Perez-Saura and Pascual Campoy},
  year          = {2024},
  eprint        = {2303.18237},
  archivePrefix = {arXiv},
  primaryClass  = {cs.RO},
  url           = {https://arxiv.org/abs/2303.18237}
}

@article{qin2018vinsmono,
  author  = {Qin, Tong and Li, Peiliang and Shen, Shaojie},
  title   = {{VINS-Mono}: A Robust and Versatile Monocular Visual-Inertial State Estimator},
  journal = {IEEE Transactions on Robotics},
  volume  = {34},
  number  = {4},
  pages   = {1004--1020},
  year    = {2018},
  doi     = {10.1109/TRO.2018.2853729}
}

@inproceedings{geneva2020openvins,
  author    = {Geneva, Patrick and Eckenhoff, Kevin and Lee, Woosik and Yang, Yulin and Huang, Guoquan},
  title     = {{OpenVINS}: A Research Platform for Visual-Inertial Estimation},
  booktitle = {Proceedings of the IEEE International Conference on Robotics and Automation (ICRA)},
  pages     = {4666--4672},
  year      = {2020},
  doi       = {10.1109/ICRA40945.2020.9196524}
}

@inproceedings{mellinger2011minimum,
  author    = {Mellinger, Daniel and Kumar, Vijay},
  title     = {Minimum Snap Trajectory Generation and Control for Quadrotors},
  booktitle = {Proceedings of the IEEE International Conference on Robotics and Automation (ICRA)},
  pages     = {2520--2525},
  year      = {2011},
  doi       = {10.1109/ICRA.2011.5980409}
}

\end{document}